\documentclass[review]{elsarticle}

\usepackage{lineno,hyperref}
\usepackage{amsmath}
\usepackage{multirow}
\modulolinenumbers[5]
\usepackage[normalem]{ulem}
\usepackage[]{color}
\useunder{\uline}{\ul}{}
\usepackage[normalem]{ulem}
\usepackage{subcaption}
\usepackage{float}
\definecolor{Orange}{rgb}{1,0.5,0}
\definecolor{Red}{rgb}{1,0,0}
\definecolor{Blue}{rgb}{0,0,1}

\journal{Information Processing \& Management}

\biboptions{authoryear}

\begin{document}

\begin{frontmatter}

\title{Local Word Vectors Guiding Keyphrase Extraction}


\author{Eirini Papagiannopoulou}
\ead{epapagia@csd.auth.gr}
\author{Grigorios Tsoumakas}
\ead{greg@csd.auth.gr}
\address{School of Informatics, Aristotle University of Thessaloniki, 54124, Greece}




\begin{abstract}
Automated keyphrase extraction is a fundamental textual information processing task concerned with the selection of representative phrases from a document that summarize its content. This work presents a novel unsupervised method for keyphrase extraction, whose main innovation is the use of {\em local} word embeddings (in particular GloVe vectors), i.e., embeddings trained from the single document under consideration. We argue that such local representation of words and keyphrases are able to accurately capture their semantics in the context of the document they are part of, and therefore can help in improving keyphrase extraction quality. Empirical results offer evidence that indeed local representations lead to better keyphrase extraction results compared to both embeddings trained on very large third corpora or larger corpora consisting of several documents of the same scientific field and to other state-of-the-art unsupervised keyphrase extraction methods.

\end{abstract}


\end{frontmatter}


\section{Introduction}
\label{introduction}

Keyphrase extraction is concerned with the selection of a set of phrases from within a document that together summarize the main topics discussed in that document \citep{hasan+ng2014}. Automatic keyphrase extraction is a fundamental task in digital content management as it can be used for document indexing, which in turns enables calculating semantic similarity between documents (and hence document clustering), and can improve browsing of digital libraries \citep{gutwin1999improving, witten1999browsing}. In addition, automatic keyphrase extraction offers an approach to document summarization. Keyphrase extraction is particularly  important in academic publishing, where it is used as a technological building block to recommend articles to readers, to highlight missing citations to authors and to analyze research trends \citep{augenstein2017semeval}. 

Supervised machine learning approaches for automatic keyphrase extraction rely on annotated corpora. However, manual selection of the keyphrases of each document by humans requires the investment of time and money and is characterized by great subjectivity. In many cases, the extracted keyphrases cover one or more non-core topics due to misunderstandings, or they miss one or more of the important topics discussed in the document. 
Using multiple annotators can partially address the problem of subjectivity by collecting more keyphrases \citep{DBLP:journals/tochi/ChuangMH12, DBLP:conf/emnlp/SterckxCDD16}. This, however, comes at the expense of additional annotation effort.
In addition, supervised methods often fail to generalize well to documents coming from a different content domain than the training corpus, may require retraining to address concept drift, and are more susceptible to varying vocabularies across documents and different personal writing styles across authors.

In contrast, this work takes a novel unsupervised path to keyphrase extraction. To be able to take into account the semantic similarity among words we consider word embeddings, in particular the one generated by GloVe \citep{Pennington14glove:global}. Different however from past approaches that exploit word embeddings in keyphrase extraction \cite{Wang2014}, we do not use pretrained vectors, but instead learn {\em local} GloVe representations in the context of {\em single documents}, in particular full-texts of academic publications. Our main hypothesis is that such local representations will be able to more accurately capture the semantic similarity of the different words and phrases in the context of each document, and help us extract more representative keyphrases, compared to global representations and other state-of-the-art unsupervised keyphrase extraction methods. Our research objective is to investigate whether this hypothesis holds. 


Our approach extracts keyphrases from the title and abstract of an academic publication, which constitute a  clear and concise summary of the whole publication, in order to avoid the noise and redundancy found in the full-text. Once local word vectors have been learned from the full-text of a given academic publication, we compute the mean vector of the words in its title and abstract, dubbed \textit{reference vector}, which we can intuitively consider as a vector representation of the semantics of the whole publication. We then extract candidate keyphrases from the title and abstract, and rank them in terms of their cosine similarity with the reference vector, assuming that the closer to the reference vector is a word vector, the more representative is the corresponding word for the publication. 



The rest of the paper is organized as follows. Section \ref{rel_work} gives a review of the related work in the field of keyphrase extraction as well as a brief overview of methods that produce word embeddings. Section \ref{method} presents the proposed approach. Section \ref{experiments} describes empirical results highlighting different aspects of our approach and comparing it with other state-of-the-art unsupervised keyphrase extraction methods. Finally, Section \ref{contributions} presents the conclusions of this work and points to future work directions.

\section{Related Work}
\label{rel_work}

\subsection{Automatic Keyphrase Extraction}
\label{previous_work}
Automatic keyphrase extraction is a  well-studied task and a variety of techniques have been proposed in the past. In this section, we present both supervised and unsupervised methods in a comprehensive and structured way.

\subsubsection{Unsupervised Approaches}
\label{rel_work_unsupevised}
Unsupervised keyphrase extraction approaches typically follow a standard three-stage process \citep{hasan+ng2010,hasan+ng2014}. The first stage concerns choosing the candidate lexical units with respect to some heuristics, such as the exclusion of stop words or the selection of words that are nouns or adjectives. The second stage concerns ranking these lexical units by measuring their {\em importance} through co-occurrence statistics or syntactic rules. The final stage concerns keyphrase formation, where the top-ranked lexical units are used either as keywords or as components of keyphrases. 

The baseline approach for unsupervised keyphrase extraction is \textit{TfIdf} \citep{DBLP:journals/jd/Jones04}. It ranks phrases in a particular document according to their frequency in this document (tf), multiplied by the inverse of their frequency in all documents of a collection (idf). Recently, \cite{DBLP:conf/ecir/FlorescuC17} proposed an approach for combining TfIdf with any other word-scoring approach. In their approach, a phrase's score is computed by multiplying its frequency within the document (tf) with the mean of the scores of the phrase's words.

Graph-based ranking algorithms are based on the following idea: first, a graph from a document is created that has as nodes the candidate keyphrases, and then edges are added between {\em related} candidate keyphrases. The final goal is the ranking of the nodes using a graph-based ranking method, such as PageRank \citep{grin+page1998}, Positional Function \citep{herings+van+talman}, and HITS \citep{kleinberg1999}. {\em TextRank} \citep{mihalcea+tatau2004} builds an undirected and unweighted graph with candidate lexical units as nodes for a specific text and adds connections (edges) between those nodes that co-occur within a window of $N$ words. The ranking algorithm runs iteratively until it converges. Once the algorithm converges, nodes are sorted by decreasing order and the top $T$ nodes form the final keyphrases. Variations of TextRank include {\em SingleRank} \citep{wan+xiao2008}, where edges have a weight equal to the number of co-occurrences of their corresponding nodes within a window, and {\em ExpandRank} \citep{wan+xiao2008}, where the graph includes as nodes not only the lexical units of a specific document but also the lexical units of the $k$ nearest neighboring documents of the initial document. In ExpandRank, an edge between two nodes exists if the corresponding words co-occur within a window of $W$ words in the whole document set. Once the graph is constructed, ExpandRank's procedure is identical to SingleRank. Recently, another unsupervised graph-based model, called PositionRank, was proposed by \cite{DBLP:conf/acl/FlorescuC17}. This method tries to capture frequent phrases taking into account, at the same time, their corresponding position in the text. More specifically, it incorporates all word's positions into a biased PageRank. Finally, the keyphrases are scored and ranked. \cite{Wang2014} propose a graph-based ranking model that takes into consideration information coming from distributed word representations. In particular, again a graph of words is initially created with edges that represent the co-existence between the words within a window of $W$ consecutive words. Then, a weight (the \textit{word attraction score}) is assigned to every edge, which is the product of two individual scores: a) the \textit{attraction force} between two words which uses the frequencies of the words as well as the distance between the corresponding word embeddings, and b) the \textit{dice coefficient} \citep{dice1945measures, Stubbs2003}. Once more, a weighted PageRank algorithm is utilized to rank the words. A similar approach that uses a personalized weighted PageRank model with pretrained word embeddings, but with different edge weights is proposed in \cite{DBLP:conf/adc/WangLM15}.

RAKE \citep{rose2010automatic} is a domain-independent and language-independent method for extracting keyphrases from individual documents. Given a list of stop words, a set of phrase delimiters, and a set of word delimiters, RAKE cuts the document text up to candidate sequences of content words and then builds a graph of word co-occurrences. Afterwards, word scores are calculated for each candidate keyword. The basic difference in comparison with the previous approaches is that RAKE is able to identify keyphrases that contain {\em interior} stop words. Specifically, RAKE detects pairs of keywords that adjoin one another at least twice in the same document, in the same order, and creates a new candidate keyphrase that contains the corresponding interior stop words.

There exists a group of approaches that incorporate knowledge from citation networks. A typical method of this group is CiteTextRank \citep{gollapalli2014extracting}, which constructs a weighted graph considering the information of short text descriptions surrounding mentions of papers (citation contexts).

Topic-based clustering methods aim at extracting keyphrases that cover all the major topics of a document. A known technique of this family is KeyCluster \citep{liu2009clustering}, which clusters similar candidate keywords utilizing Wikipedia and co-occurrence statistics. The basic idea is that each cluster corresponds to a specific topic of the document and by selecting candidate keyphrases from each cluster, all the topics are covered. TopicRank \citep{bougouin2013topicrank} is another method that extracts keyphrases from the most significant topics of a document. First, the text of interest is preprocessed and then keyphrase candidates are grouped into separate topics using hierarchical agglomerative clustering. In the next stage, a graph of topics is constructed whose edges are weighted based on a measure that considers phrases' offset positions in the text. As a final step, TextRank is used to rank the topics. Topical PageRank (TPR) \citep{liu2010automatic} is an alternative methodology which first obtains the topics of words and documents using Latent Dirichlet Allocation (LDA) \citep{blei2003latent} and then begins the construction of the word graph for a given document. The idea of TPR is to run a PageRank for each topic separately by modifying the basic PageRank score function utilizing the word topic distributions calculated earlier for the given document.

\cite{Tomokiyo:2003:LMA:1119282.1119287} create both unigram and n-gram language models on a foreground corpus (target document) and a background corpus (document set). Their main idea is based on the fact that the loss between two language models can be measured using the Kullback-Leibler divergence. Particularly, in a phrase level, for each phrase, they compute the \textit{phraseness} as the divergence between the unigram and n-gram language models on the foreground corpus and the \textit{informativeness} as the divergence between the n-gram language models on the foreground and the background corpus. Finally, they sum the phraseness and informativeness to obtain a final score for each phrase and sort them by this score.

\subsubsection{Supervised Approaches}
\label{rel_work_supevised}
In supervised learning, a classifier is trained on annotated with keyphrases documents in order to determine whether a candidate phrase is a keyphrase or not. These keyphrases and non-keyphrases are used to generate positive and negative examples.

The famous KEA system \citep{witten1999kea} is one of the first supervised keyphrase extraction systems which uses only two features during training and extraction process: TfIdf and first occurrence attribute. The training stage uses documents whose the keyphrases are known. Then, for each document, candidate keyphrases are identified and their feature values are calculated. Finally, KEA uses an expression to rank the candidates, that incorporates the corresponding features, based on Naive Bayes. Later, another system which uses linguistic knowledge has been proposed by \cite{Hulth:2003:IAK:1119355.1119383}. For each candidate phrase of the training data, that has been selected in an earlier stage, four features are calculated: the within-document frequency, the collection frequency, the relative position of the first occurrence, and POS tag(s). Finally, the machine learning approach is a rule induction system with bagging. The popular keyphrase extraction system, called Maui \citep{medelyan2009human}, first determines all n-grams up to 3 words and then calculates a set of meaningful features such as TfIdf, the position of the first occurrence, \textit{keyphraseness}, phrase length, and features based on Wikipedia statistics which are used in its classification model. In \cite{caragea2014citation}, a binary classification model, CeKE, has been proposed (Naive Bayes classifier with decision threshold 0.9) which utilizes \textit{novel} features from the information of citation contexts and existing features from previous works. Recently, \cite{DBLP:conf/emnlp/SterckxCDD16} conduct an interesting study where they conclude that unlabeled keyphrase candidates are not reliable as negative examples. For this reason, they propose to treat supervised keyphrase extraction as Positive Unlabeled Learning by assigning weights to training examples, modeling in this way the uncertainty. Firstly, they train a classifier on a single annotator's data and use the predictions on the negative/unlabeled phrases as weights. Then, another classifier is trained on the weighted data mentioned above in order to predict the labels of the candidates.

Other approaches that use neural network models have been proposed, with the most recent work to be a generative model for keyphrase prediction using an encoder-decoder framework that tries to capture the semantic meaning of the content via a deep learning method \citep{P17-1054}. In fact, it applies a recurrent neural network (RNN) Encoder-Decoder model in order to learn the mapping from the source text to its corresponding target keyphrases. The main drawback with such approaches is that the model is expected to work well on text documents that have the same domain with the training data.

Another point of view is to see keyphrase extraction as a learning to rank task such as in \cite{jiang2009ranking}. The basic reason to adopt this approach is the fact that it is easier to determine if a candidate phrase is a keyphrase in comparison with another candidate phrase than to classify it as a keyphrase or not, by taking such hard decisions.

We should not omit the recent work on keyphrase extraction where the task has been treated as a sequence tagging task using Conditional Random Fields \citep{DBLP:conf/aaai/GollapalliLY17}. The features used represent linguistic, orthographic, and structure information from the document. Furthermore, they investigate feature-labeling and posterior regularization in order to integrate expert/domain-knowledge throughout the keyphrase extraction process.

\subsection{Dense Vectors}
Since 1990, a great number of methods have been proposed for words' representation, such as the popular Latent Dirichlet Allocation (LDA) \citep{blei2003latent} and Latent Semantic Analysis (LSA) \citep{deerwester1989computer,deerwester1990indexing}. Generally, such approaches that are based on co-occurrences' matrix \citep{deerwester1990indexing, lund1996producing, blei2003latent} are able to capture semantics and are also used for further dimensionality reduction. However, \cite{Bengio:2003:NPL:944919.944966} invented the term ``word embeddings'', proposing a simple feed-forward neural network which predicts the next word in a sequence of words. In fact, word embeddings came to the foreground by \cite{DBLP:journals/corr/abs-1301-3781}, who presented the well-known Continuous Bag-of-Words Model (CBOW) and the Continuous Skip-gram Model, establishing widely the use of pretrained embeddings.

In this work, we utilize the GloVe (Global Vectors) \citep{Pennington14glove:global} method for the generation of the word vectors. This methodology exploits statistical information by training only on the non-zero elements in a word-word co-occurrence matrix in an efficient way and finally, creates a meaningful word vector space.

\section{The Reference Vector Algorithm}
\label{method}

\begin{figure}[h]
\centering
  \centering
  \includegraphics[width=12cm,height=34cm,keepaspectratio]{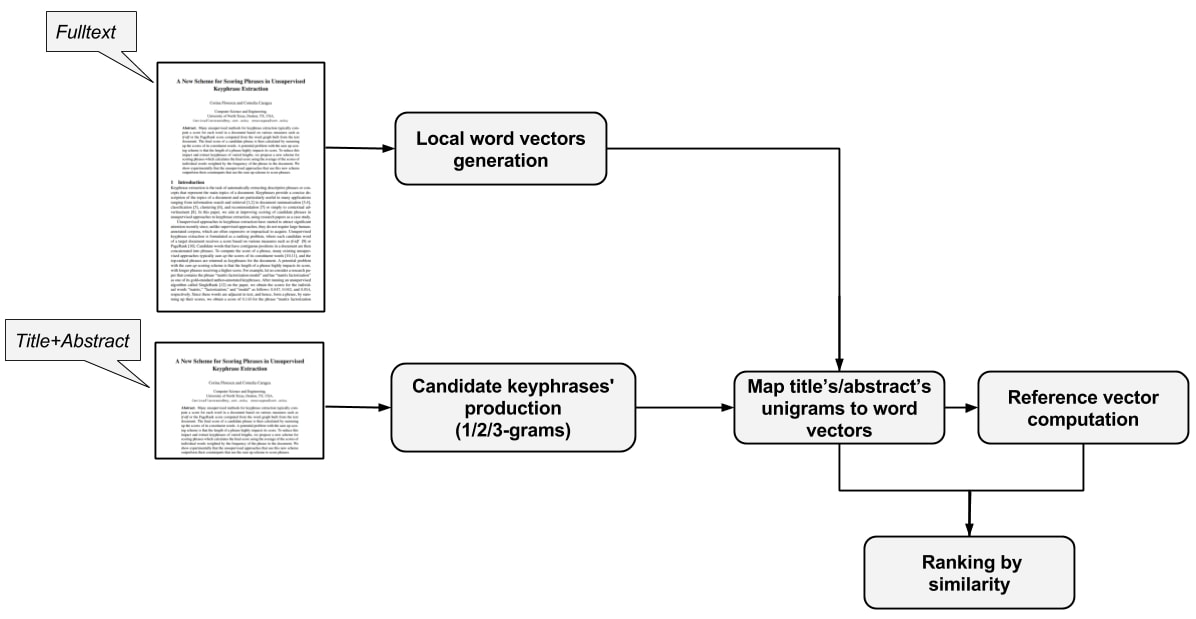}
  \caption{System processing pipeline.} 
  \label{fig:system-pipeline}
\end{figure}

This section describes thoroughly our approach, called Reference Vector Algorithm (RVA), for extracting keyphrases from the titles and abstracts of scientific articles. Our approach exploits the GloVe word vector representation to detect the candidate keywords and to provide a complete set of representative keyphrases for a particular title and abstract. Fig. \ref{fig:system-pipeline} summarizes the processing pipeline of RVA.




\subsection{Candidate Keyphrases' Production}

We follow the choice of previous keyphrase extraction systems \citep{Hulth:2003:IAK:1119355.1119383, medelyan2009human} by extracting only unigrams, bigrams, and trigrams, as these are the most frequent lengths of keyphrases that are met in the datasets used in the experimental study (see Section \ref{datasets} for more details about the datasets' statistics). In this way, we can effectively reduce the number of possible $n$-grams that are candidates as keyphrases by restricting the value of $n$ to $\{1, 2, 3\}$, with respect to the observation that, in general, a document's keyphrases tend to be up to trigrams \citep{DBLP:conf/aaai/GollapalliLY17}.

\textbf{Candidate Unigrams}: Unigrams constitute the smallest but the most significant parts that form the longer keyphrases. The criteria for the selection of the appropriate unigrams are the following:
\begin{itemize}
\item candidates should have word length lower than 36 and greater than 2 characters (a quite wide range, as the longest word in the well-known Oxford English Dictionary contains 30 characters),
\item they do not belong to the stop words list defined by us,
\item they are not numbers,
\item they do not include the following set of characters: !, @, \#, \$, *, =, +, ., ‘,’, ?, $>$, $<$, \&, (, ), \{, \}, [, ], $\mid$

\end{itemize}


\textbf{Candidate Bigrams}: We choose as candidate bigrams those whose words are in candidate unigrams and appear in the text in that specific sequence. We do not keep as candidate bigrams those whose the length of both words is lower than 4. 

\textbf{Candidate Trigrams}: We apply the same procedure as above (for bigrams).

\subsection{Scoring the Candidate Keyphrases}

As a first step, we produce the local word vectors by applying the GloVe model to the target full-text scientific publication (that's why we call them {\em local} vectors). 
The word vectors generated by GloVe, in the context of only one article, encode the role of words as expressive means of writing, via a vector representation. This type of representation can capture how a limited vocabulary is structured and extended within the narrow limits of a scientific paper. We choose the GloVe technique, instead of other word vector representations, as it is based on the full-text co-occurrence statistics within a predetermined window. GloVe builds an overview of the neighborhood of each one word and, simultaneously, provides us with a picture of its \textit{local contexts}. For the purposes discussed above, we utilize the implementation of the GloVe model for learning word vector representations that is publicly available by Stanford University on GitHub \footnote{\url{https://github.com/stanfordnlp/GloVe}}. 


We compute the mean vector of the title and the abstract, called \textit{reference vector} by averaging the individual local word vectors that appear in that text segment. First, we sum all the word vectors which match to the candidate unigrams formed in the previous step. Then, the reference vector is derived by dividing with the number of candidate unigrams contained in the title and the abstract. We should take into account that the more often a word shows up in the target-text, the more it affects the reference vector. Finally, we calculate the cosine similarity between each candidate unigram's local vector that appears in the text segment and the reference vector, creating a \textit{mapping} between the words and their corresponding cosine similarity scores.

As a scoring function for a candidate bigram or trigram, we choose the sum of the individual words' scores, as we prefer the informativeness come from the longer keyphrases rather than the shorter ones e.g. the unigrams. A great number of existing unsupervised approaches sum up the individual word scores to produce the final phrase score \citep{mihalcea+tatau2004, wan+xiao2008}. In this way, we expand the \textit{mapping} mentioned above with the bigrams/trigrams and their corresponding score.

Note that most approaches include the Part-of-Speech (PoS) tagging stage based on the observation that the lexical units which belong to a keyphrase often are nouns, adjectives or adverbs (see Sections \ref{rel_work_unsupevised}, \ref{rel_work_supevised}). In addition to PoS tagging, stemming is another basic preprocessing step that is suggested by some approaches such as in \cite{Hulth:2003:IAK:1119355.1119383}. Our decision, to use the word vectors' representation mentioned above, provides us with the advantage to avoid such additional and time-consuming processes, as GloVe is designed to capture in a quantitative way the nuance necessary to discriminate two individual words by associating more than a single number to them, utilizing the vector difference between the two corresponding word vectors.   

\section{Experiments}
\label{experiments}

We first present the two collections that were used in our empirical study along with some interesting statistics. Then, we describe the evaluation framework and the experimental setup. Finally, we discuss in detail the results, providing both a quantitative and a qualitative evaluation of the proposed approach. 

\subsection{Data Sets and their Statistics}
\label{datasets}

Our empirical study is based on 2 popular collections of scientific publications: a) Krapivin \citep{krapivin2009}, which contains 2304 scientific full-text articles from computer science domain published by ACM, along with author-assigned and editor-corrected keyphrases, and b) Semeval \citep{Kim:semeval2010}, which contains 244 scientific full-text articles from
the ACM Digital Library, along with author-assigned, as well as reader-assigned keyphrases. We apply a preprocessing stage on both datasets in order to separate the upper part (title, abstract for Krapivin and title, abstract, Categories/Subject Descriptors as well as General Terms of the ACM's Computing Classification System for Semeval) of each document from the remaining part (main text body). The refinement process of the Krapivin dataset was quite simple as the title and the abstract are clearly indicated. However, the corresponding separation process for the Semeval dataset was based on heuristic rules (the main body usually starts with a section that contains in its title derivatives of the word ``introduction'' and it is located below the Categories/Subject Descriptors and the General Terms).



\begin{figure}[h]
\centering
  \centering
  \includegraphics[width=1.0\linewidth]{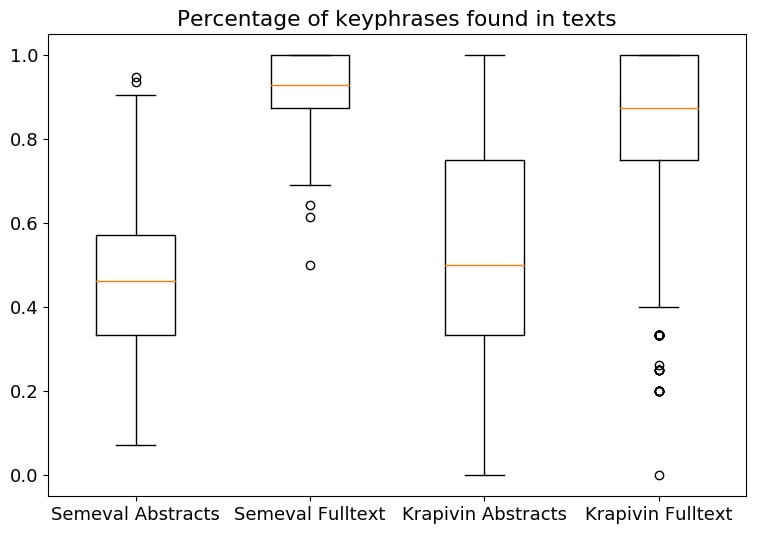}
  \caption{Box plots of the percentage of keyphrases found in the abstract and full-text of each of the two collections. The two box plots on the left correspond to the {\em Semeval} collection, while the two on the right correspond to the {\em Krapivin} collection.} 
  \label{fig:keyphrase_percentage}
\end{figure}

Figure \ref{fig:keyphrase_percentage} presents box and whisker plots of the percentage of the {\em gold}, i.e., ground truth keyphrases appearing in the abstract, and in the full-text of each scientific publication of both collections. We notice that full-texts include the great majority of the gold keyphrases with a mean value approximately equal to 90\%. Abstracts, on the other hand, include approximately half of the gold keyphrases on average. 


\begin{figure}[h]
  \centering
  \includegraphics[width=1.0\linewidth]{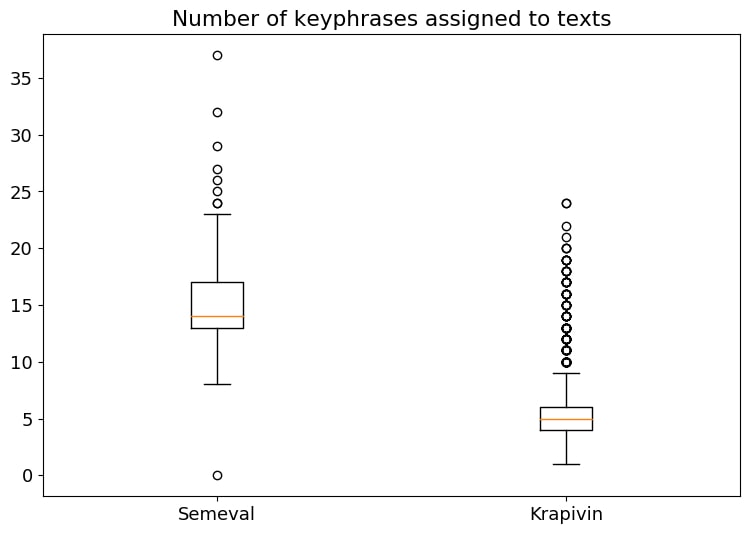}
   \caption{Box plots of the number of keyphrases for Semeval (left) and Krapivin (right).} 
  \label{fig:keyphrase_number}
\end{figure}

Figure \ref{fig:keyphrase_number} presents box and whisker plots of the number of gold keyphrases that are associated with each article of the two collections. We can see that in {\em Semeval} an average number of 14 keyphrases is assigned per document, whereas the main range of values is from 13 to 17 keyphrases. In the case of {\em Krapivin}, the main range of values varies from 4 to 6, with a mean value approximately equal to 5. 

\begin{figure}[h]
  \centering
  \includegraphics[width=1.0\linewidth]{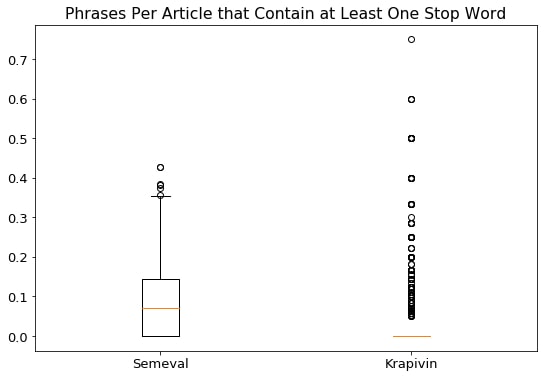}
   \caption{Box plots of the percentage of ``gold'' keyphrases per article that include at least one stop word for Semeval (left) and Krapivin (right).} 
  \label{fig:stopwords}
\end{figure}

Figure \ref{fig:stopwords} presents box and whisker plots of the percentage of ``gold'' keyphrases per article that include at least one stop word. The keyphrases of {\em Semeval} dataset have quite high percentages of phrases with stop words in comparison with the {\em Krapivin} dataset.

Finally, Table \ref{length} presents the frequency of the gold keyphrases in each collection per different length (number of words). We can see that most keyphrases are bigrams, followed by unigrams/trigrams. Keyphrases with 4 to 6 words are less frequent, while there also exist a couple of outliers with 7 to 9 words. 

\begin{table}[h]
\centering
\begin{tabular}{|l|lllllllll|}
\hline
Data sets          & 1    & 2    & 3    & 4   & 5  & 6  & 7 & 8 & 9  \\\hline
Semeval           & 759  & 2005 & 782  & 171 & 46 & 16 & 3 & 2 & 1   \\
Krapivin          & 2330 & 7575 & 1936 & 364 & 70 & 18 & 2 & 1 & 0   \\\hline
\end{tabular}
\caption{Frequency of gold keyphrases per length (number of words), ranging from 1 to 9.}
\label{length}
\end{table}

\subsection{Experimental Setup}


It is not clear how to properly evaluate a returned n-gram phrase given a golden multiword keyphrase, especially in cases where the returned n-gram is part of or longer than the golden keyphrase. For this reason, we follow the evaluation process of \cite{DBLP:conf/ecir/RousseauV15}, which calculates the $F_1$-measure between the set of words found in all golden keyphrases and the set of words found in all extracted keyphrases. 

Regarding the GloVe setup, we used the default parameters ($x_{max}$ = 100, $\alpha = \frac{3}{4}$, \textit{window size} = 10), as they are set in the experiments of \cite{Pennington14glove:global} who used empirically parameter tuning to find the best values. 
Finally, we produce 50 and 200-dimensional vectors with 50 iterations as indicated in \cite{Pennington14glove:global} for vectors smaller than 300 dimensions. Such dimensions are appropriate as there are also pretrained GloVe word embeddings for 50 and 200-dimensional vectors, useful for the comparison between local and pretrained word vectors.

We propose the number of keyphrases to be determined based on the title's and abstract's size. Particularly, we choose a more flexible threshold for the number of the representative phrases that will be returned as keyphrases which is inspired by \cite{mihalcea+tatau2004}, i.e., the selection of the top-scored $N$ phrases as keyphrases, where $N$ is equal to $\frac{1}{3}$ of the number of the different words in the title and abstract, rather than to set a fixed number, which would be a quite strict decision considering Fig. \ref{fig:keyphrase_number}. 

We utilize the PKE, which is an open source python-based keyphrase extraction toolkit \citep{DBLP:conf/coling/Boudin16}, for the experiments with TfIdf and the graph-based approaches. The code for the RVA method will be uploaded to our Github repository \footnote{\url{https://github.com/epapagia/RVA}}. The
datasets with the abstracts of Krapivin and Semeval are available for research purposes \footnote{\url{https://github.com/epapagia/Datasets-Keyphrase-Extraction}}

\subsection{RVA Variants Evaluation Based on Text Size for GloVe Training}
\label{rva_variants}

In this section, we give a general view of the performance of the RVA algorithm by changing the dimension of the word vectors and training the GloVe model on different corpus sizes, including the use of word vectors trained on massive web datasets (pretrained word vectors). Specifically, we ran experiments with the pretrained word vectors that were created by training on Wikipedia 2014 + Gigaword 5 which have 400000 vocabulary size, uncased. Furthermore, according to the RVA's methodology, we generated local word vectors from each one scientific publication of the 2 collections, keeping them in separate files. Finally, we trained a GloVe model on the smaller dataset collections of Semeval and Krapivin, separately, as an intermediate size of the corpus. In all cases, except for those of the pretrained word vectors, we have included in the vocabulary all the possible words that appear in the texts.

As this is the first time that such local word vectors are utilized in this way, we prepared an experimental study appropriate to provide us with comprehensive conclusions about the functionality of the reference vectors as guides of the keyphrase extraction process. For this reason, we have designed 2 additional different versions of the proposed RVA approach called \textit{Full-text Reference Vector Algorithm (RVA-F-F)} and \textit{Reference Vector Algorithm using Full-text Candidates (RVA-A-F)}, respectively. The RVA-F-F follows exactly the same process as the RVA. The main difference is that the reference vector, i.e., the mean vector, is computed by averaging the individual local word vectors that appear in the full-text, not only in the title and the abstract. The candidate words are, also, extracted from the full-text article. On the other hand, the main difference between RVA and RVA-A-F is that the latter uses candidate unigrams, bigrams and trigrams from the whole article without being limited to the article's summary like RVA. However, the reference vector is still calculated based only on the title and the abstract. Conventionally, the first letter after the first dash indicates the part of the text from where the reference vector is calculated (A for abstract and title and F for full-text), whereas the second letter refers to the part of the text from which the candidate keywords come. For consistency reasons, the proposed method is denoted as RVA-A-A. Table \ref{tbl:setup} describes in detail the different word vector settings with respect to the vector dimensions and the text size used for training of GloVe, providing the corresponding abbreviations that are used in the results' Table \ref{RVA-variants}.

\begin{table}[h]
\centering
\small
\begin{tabular}{|l|l|}
\hline
Abbreviation  &  Description \\\hline
LOC-50        & local 50-dim vectors - trained on individual files   \\
LOC-200       & local 200-dim vectors - trained on individual files  \\
CV-50    &  50-dim vectors - trained on each collection separately   \\
CV-200    & 200-dim vectors - trained on each collection separately   \\
PV-50        & 50-dim pretrained vectors  \\
PV-200       & 200-dim pretrained vectors  \\\hline
\end{tabular}
\caption{Explanation of the abbreviations used in the tables with the experimental results to describe the  settings of the proposed method and its variants.}
\label{tbl:setup}
\end{table}

Experiments conducted for word vectors with dimensions 50 and 200 for all RVA variants (RVA-A-A, RVA-F-F, RVA-A-F) and the corresponding results show that the vectors' dimensions do not substantially affect the methods' performance. 
Regarding the size of the text on which the method was trained to produce the respective word vectors, we see that the usage of local word vectors in all RVA variants outperforms the experimental results where collection word vectors or pretrained word vectors are used. More specifically, in Semeval RVA-A-A with local word vectors achieves approximately 0.37 in the $F_1$ score, whereas with the utilization of collection word vectors or pretrained ones performs worse, with 0.34 and 0.30 scores, respectively. For the large dataset of Krapivin, the results of RVA-A-A have again the same ranking, i.e., the settings with local word vectors are the winners (0.32), followed by the setup of collection word vectors (0.28). The cases with pretrained word vectors are once more last in the ranking (0.26 $F_1$-measure). The results of RVA-F-F and RVA-A-F follow the same ordering.  However, the differences in the usage of local, collection and pretrained word vectors range at higher levels for RVA-F-F and lower for RVA-A-F.

\begin{table}[h]
\centering
\scalebox{0.90}{
\begin{tabular}{|l|c|c|c|c|c|c|}
\hline
\multicolumn{7}{|c|}{$F_1$-measure}                                                                            \\ \hline
\multicolumn{1}{|c|}{\multirow{2}{*}{Setup}} & \multicolumn{3}{c|}{Semeval} & \multicolumn{3}{c|}{Krapivin} \\ \cline{2-7} 
\multicolumn{1}{|c|}{}                       & RVA-A-A  & RVA-F-F & RVA-A-F & RVA-A-A  & RVA-F-F  & RVA-A-F \\ \hline
LOC-50                                       & \textbf{0.36815}  & 0.29543 & 0.11353 & \textbf{0.32062}  & 0.21171  & 0.06265 \\ 
LOC-200                                      & 0.36493  & 0.29641 & 0.11259 & 0.31999  & 0.20984  & 0.06212 \\ 
CV-50                                        & 0.34202  & 0.23608 & 0.06779 & 0.28149  & 0.14415  & 0.03893 \\ 
CV-200                                       & 0.34122  & 0.23535 & 0.06904 & 0.28267  & 0.15075  & 0.03944 \\ 
PV-50                                        & 0.30188  & 0.15535 & 0.02026 & 0.25903  & 0.09089  & 0.01488 \\ 
PV-200                                       & 0.30015  & 0.15505 & 0.02209 & 0.25804  & 0.09078  & 0.01583  \\ \hline
\end{tabular}}
\caption{Experimental results for RVA-A-A, RVA-F-F and RVA-A-F using different settings for the GloVe method. The $2^{nd}, 3^{rd}$ and $4^{th}$ columns concern the Semeval dataset, whereas the results of the last 3 columns correspond to the articles of the Krapivin dataset.}
\label{RVA-variants}
\end{table}


Regarding the use of local word vectors instead of the pretrained ones, the experimental results essentially confirm that a small text (e.g. in the size of a scientific publication that refers to a specific subject) is possible to offer a sufficient basis to GloVe for the keyphrase extraction task. That is due to the fact that a scientific article includes a complete textual description of a specific topic, usually structured, using a limited but adequate vocabulary to reflect the semantic context of its words. Despite the fact that the pretrained word vectors succeed in capturing more general meanings as well as the underlying concepts that distinguish the words, performing well in tasks like the one of word analogy, we note that in the task of keyphrase extraction, they have low performance. Apparently, the generalization that is included in the pretrained word vectors is unnecessary and incorporates probably ``noise'' (redundant information). The above claim is also corroborated by the results when collection word vectors are used; the more local the word vectors, the better are the results.

At this point, we focus on the first two columns of each dataset trying to explain why we prefer the title and the abstract as the most suitable part of the article for the calculation of the reference vector rather than the full-text. As we can see, RVA-A-A clearly outperforms RVA-F-F in all cases. Indicatively, we mention that for the Semeval the RVA-A-A LOC-50 achieves approximately 0.37 while for the RVA-F-F LOC-50 the $F_1$ score equals to 0.30. For the Krapivin dataset, we have exactly the same ranking, however, with a higher difference between the two RVA variants (greater than 0.10). The intuition is that, possibly, there are frequent words in the full-text which play an important role in the text structure and in the content's presentation. However, these words have an auxiliary role and they could not be considered as keywords. Unfortunately, when we extract candidates from the whole article, the importance of such words is reflected in the computation of the reference vector. For this reason, we conclude that it is much safer to use only the title and the abstract instead of the full-text, or generally speaking, parts of a limited text that contain some semantically significant and meaningful words, avoiding in this way the noise of the full-text.

The final issue that we discuss in this section is whether the candidate keywords should come from abstracts or from full-texts regardless of the calculation of the reference vector, i.e., keeping as reference vector the mean word vector of the title and the abstract. Focusing on the $1^{st}$ and the $3^{rd}$ column of each dataset, we notice that there is a dramatic reduction in the rates in all the cases of RVA-A-F where we take into account as candidate keywords, the words from the full-text article (greater than -0.24). The above results lead us to the conclusion that it is more efficient for all the candidate words to participate in the calculation of the reference vector, as this facilitates the keywords' detection.

\subsection{Comparison with Other Methods}

We compare the standard version of RVA (GloVe vectors of dimension 50 trained on each full-text) to TfIdf and 4 graph-based approaches, namely SingleRank \citep{wan+xiao2008}, TopicRank \citep{bougouin2013topicrank}, WordAttractionRank (WARank) \citep{Wang2014} and its extended version (WARank2015) \citep{DBLP:conf/adc/WangLM15}. TfIdf, SingleRank and TopicRank are considered state-of-the-art methods for keyphrase extraction \citep{DBLP:journals/lre/KimMKB13,hasan+ng2014}. The \textit{document frequency} used by TfIdf approach is calculated separately for each dataset collection. Graph-based methods are employed using their default parameters as finally set in the corresponding papers. For WARank and WARank2015, we used the pretrained word embeddings from \citep{DBLP:journals/jmlr/CollobertWBKKK11}, which were also used in \citep{Wang2014} and \citep{DBLP:conf/adc/WangLM15}. We experimented with two versions of each competitor of RVA, one using the abstracts and one using the full-texts of each article. 


\subsubsection{Results}
\label{exp_results}

Table \ref{final_results} shows the $F_1$ score of each method in each of the two datasets, sorted in descending order. We first notice that in both datasets the full-text version of TfIdf is much better than the abstract version. This is no surprise, as abstracts do not contain enough text to enable the separation of keyphrases from non-keyphrases in contrast with the full-text of articles. For the 4 graph-based methods, the opposite is observed, similarly to what we noticed for RVA in Section \ref{rva_variants}. It appears that, despite their smaller size, abstracts capture adequately the co-occurrence (proximity) of words that is necessary for graph creation, avoiding at the same time the noise (lots of unimportant words) in the full-texts. We focus on the best versions of TfIdf and the 4 graph-based approaches in the rest of this section. 


We then notice that RVA achieves the best place in both datasets. SingleRank-ab is 2nd in Krapivin and 3rd in Semeval. TfIdf-ft is 2nd in Semeval (without large difference from RVA) and 3rd in Krapivin (without large difference from SingleRank-ab). The other 3 graph-based methods follow in positions 4 to 6, without large differences among them. WARank-ab is better than WARank2015-ab in both datasets. We therefore focus on WARank-ab only in the rest of this section. 


\begin{table}[h]
\centering
\begin{tabular}{|l|l|l|l|}
\hline
Method        & Semeval         & Method        & Krapivin         \\ \hline
RVA           & \textbf{0.36815}& RVA           &\textbf{0.32062} \\ 
TfIdf-ft      & 0.36114    		& SingleRank-ab & 0.27795    \\ 
SingleRank-ab & 0.33043         & TfIdf-ft      & 0.27668          \\ 
WARank-ab	  & 0.32797 		& WARank-ab	    & 0.27436    \\ 
TopicRank-ab  & 0.32571         & WARank2015-ab  & 0.27365          \\ 
WARank2015-ab  & 0.32553    	& TopicRank-ab & 0.27038    \\ 
TopicRank-ft   & 0.32044 		& TfIdf-ab      & 0.23196  \\
SingleRank-ft & 0.28401 		& SingleRank-ft & 0.23088 \\
WARank2015-ft & 0.27799 		& TopicRank-ft  & 0.23032 \\
TfIdf-ab      & 0.26102         & WARank2015-ft & 0.18934          \\ 
WARank-ft 	  & 0.22005 		& WARank-ft 	& 0.16869 \\
\hline
\end{tabular}
\caption{Experimental results ($F_1$ measure) for the baseline TfIdf and the methods TopicRank, SingleRank, WARank, and WARank2015 as well as the proposed method RVA. The ``-ab'' at the end of the methods' names implies that they are applied only on titles and abstracts, whereas the ``-ft'' means that keyphrases are extracted from the full-text of the articles. Methods are ordered in descending $F_1$ measure.}
\label{final_results}
\end{table}


We further employ statistical tests to compare RVA against each one of TfIdf-ft, SingleRank-ab, TopicRank-ab and WARank-ab. For each dataset and pair of methods, we test if the differences of the $F_1$ scores across articles are statistically significant. For Semeval, we used the paired-t-test, as 
the differences in the $F_1$ score of RVA and each other method in each article are approximately normally distributed according to the Shapiro-Wilk test at the 0.01 significance level (Table \ref{normality_tests}). For Krapivin, we used the Wilcoxon test, as the normality assumption on the differences of the $F_1$ scores is rejected by the Shapiro-Wilk test. Table \ref{significance_tests} shows that RVA is significantly better than the competing methods in both datasets with the exception of TfIdf-ft in Semeval. Note however, that Semeval is a much smaller dataset (244 articles) compared to Krapivin (2304 articles). 

A possible reason for the quite high performance of TfIdf-ft in Semeval is the fact that a considerable ratio of its articles' ``gold'' keyphrases include stop words (see Fig. \ref{fig:stopwords} in Section \ref{datasets}). RVA does not return phrases with stop words, while TfIdf-ft extracts all possible n-grams ($n$ $\in$ $\{1, 2, 3\}$) without excluding stop words. 

\begin{table}[h]
\centering
\begin{tabular}{|l|c|c|}
\hline
\multicolumn{3}{|c|}{Shapiro-Wilk (p-values)}                                        \\ \hline
Method                & \multicolumn{1}{l|}{Semeval} & \multicolumn{1}{l|}{Krapivin} \\ \hline
TopicRank-ab          & 0.153                        & $\approx$0.000                         \\
SingleRank-ab         & 0.129                        & $\approx$0.000                         \\
TfIdf-ft              & 0.221                        & $\approx$0.000                         \\
WARank-ab & 0.113                        & $\approx$0.000                         \\ \hline
\end{tabular}
\caption{Results from the Shapiro-Wilk test on the performance differences between RVA and the other methods.}
\label{normality_tests}
\end{table}

\begin{table}[h]
\centering
\begin{tabular}{|l|c|c|}
\hline
\multicolumn{3}{|c|}{P-values}                                                                                  \\ \hline
Method                & \multicolumn{1}{l|}{Semeval (Paired-t-test)} & \multicolumn{1}{l|}{Krapivin (Wilcoxon)} \\ \hline
TopicRank-ab          & $\approx$0.000                                        & $\approx$0.000                                    \\
SingleRank-ab         & $\approx$0.000                                        & $\approx$0.000                                    \\
TfIdf-ft              & 0.240                                        & $\approx$0.000                                    \\
WARank-ab & $\approx$0.000                                        & $\approx$0.000                                    \\ \hline
\end{tabular}
\caption{Paired-t-test and Wilcoxon test results between RVA and the other methods. Each row of the table contains the comparison with a specific method.}
\label{significance_tests}
\end{table}


\subsubsection{Discussion}
\label{discussion}

The graph-based methods with which we have experimented, first, construct a graph of words (SingleRank, WARank) or topics (TopicRank) based on the position of the words in the text. For example, SingleRank assigns weights to the graph edges utilizing the co-occurrence of words in a given window; TopicRank uses distances between words' offset positions in the document; WARank exploits information that incorporates word frequencies as well as semantic distances between pretrained word embeddings. 
Then, the PageRank algorithm determines the final score of the words/topics (graph vertices), recursively, using information coming from the graph's links. On the other hand, GloVe is an unsupervised method that produces word vector representations. Its aim is to learn word vectors such that their dot product is equal to the logarithm of the words' probability of co-occurrence. Both graph-based approaches and GloVe capture information from the neighborhood of each one word (text statistics). However, in the context of our method, GloVe produces local word vectors, which is a \textit{more expressive representation} than the assignment of a simple number as a score to each word by PageRank. 
This word vector representation allows us to express the article's summary (title and abstract) with only one vector, which then guides the ranking of the candidate words as keywords. Moreover, the strong baseline TfIdf focuses on each word separately without taking into account any information related to the context of the words, but only to their frequency. As a result, TfIdf cannot capture any interrelationships between words in the text as well as word semantics.

Moreover, we propose an alternative (\textit{semantic}) evaluation, focused on the ``gold'' keyphrases' comparison with the returned keyphrases of the best systems given above, that exploits the representation of the words as vectors. Particularly, considering for each article as the ``gold'' reference vector the mean word vector derived from the ground truth's keyphrases, we compute the cosine similarity between this vector and the following:
\begin{itemize}
\item[i.] the mean vector of the summary (i.e. title and abstract) which is used as a reference vector by RVA,
\item[ii.] the mean vector of the suggested keyphrases by RVA,
\item[iii.] the mean vector of the rejected candidates by RVA, as they appear in a low ranking,
\item[iv.] the mean vectors that come from the proposed keyphrases by the four best systems (i.e. TfIdf-ft, SingleRank-ab, WARank-ab, and TopicRank-ab), using the same local word vectors that are produced in the context of RVA.
\end{itemize}

\begin{figure}[H]
\centering
\begin{subfigure}[b]{1\textwidth}
   \includegraphics[width=1\linewidth]{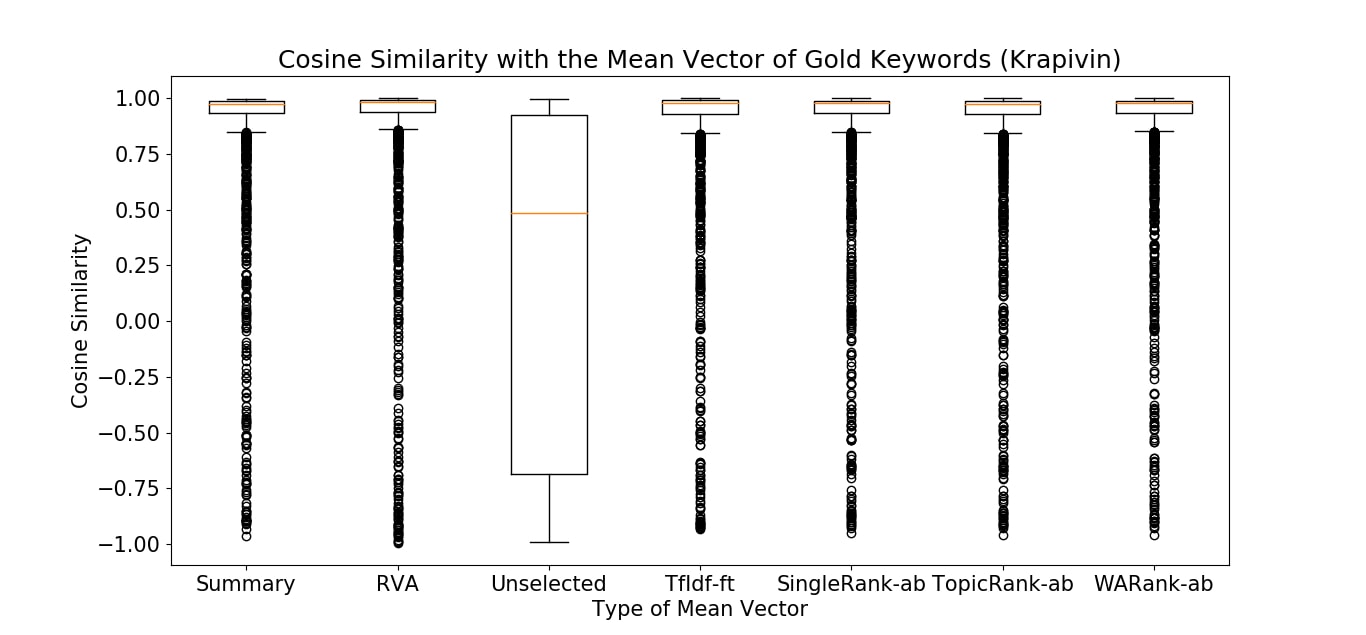}
   \caption{}
   \label{fig:eval-based-on-vectors-krpvn} 
\end{subfigure}

\begin{subfigure}[b]{1\textwidth}
   \includegraphics[width=1\linewidth]{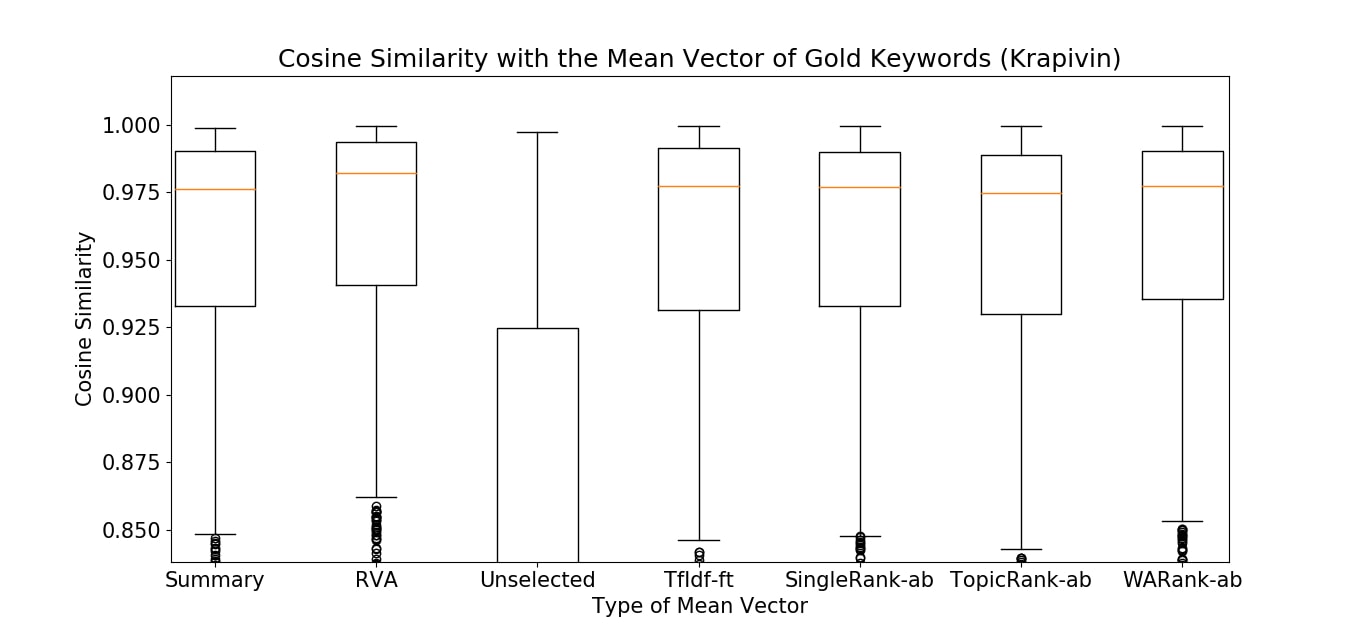}
   \caption{}
   \label{fig:eval-based-on-vectors-krpvn-zoom}
\end{subfigure}
\caption{Semantic evaluation on the Krapivin dataset. (a) Quality evaluation of methods based on the word vector representation. (b) Close-up view.}
\label{fig:semantic-eval-Krap}
\end{figure}

\begin{figure}[H]
\centering
\begin{subfigure}[b]{1\textwidth}
   \includegraphics[width=1\linewidth]{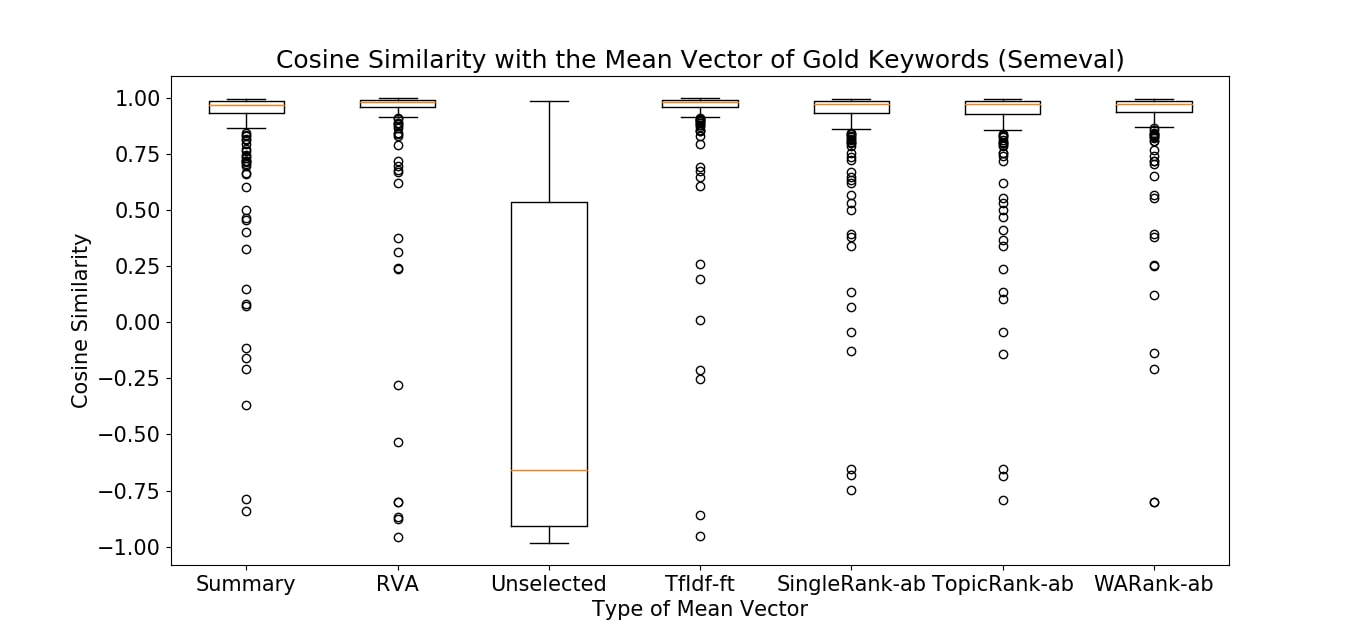}
   \caption{}
   \label{fig:eval-based-on-vectors-smvl} 
\end{subfigure}

\begin{subfigure}[b]{1\textwidth}
   \includegraphics[width=1\linewidth]{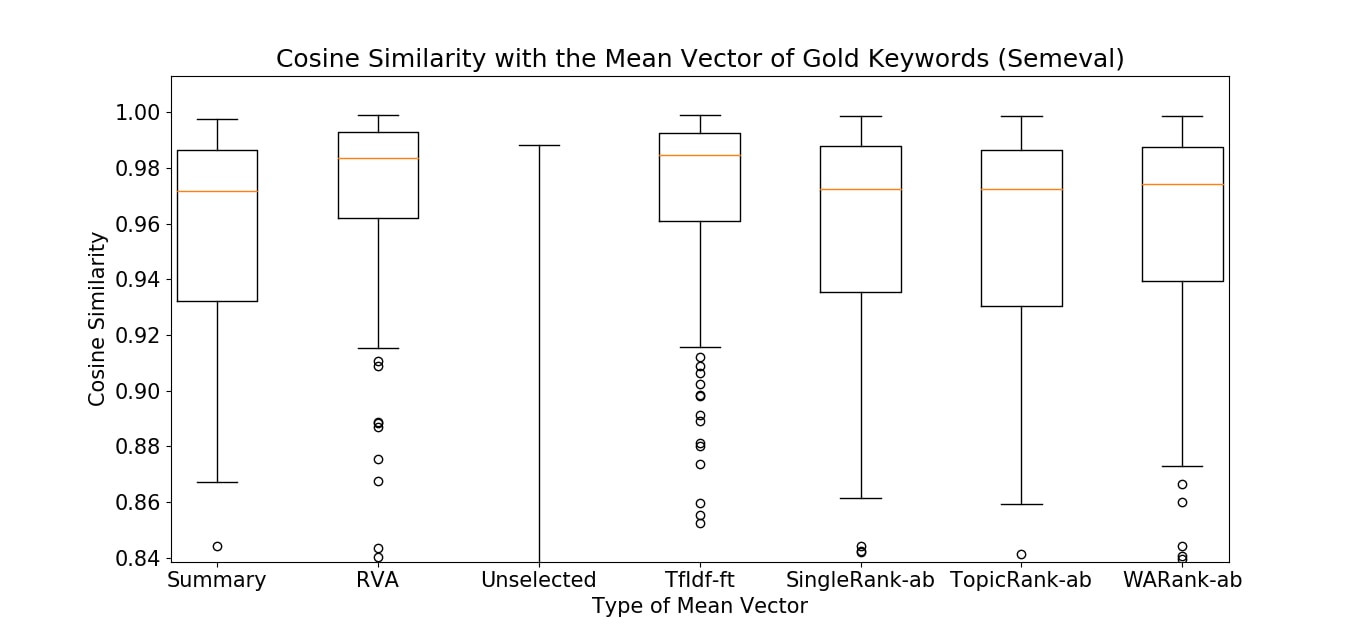}
   \caption{}
   \label{fig:eval-based-on-vectors-smvl-zoom}
\end{subfigure}
\caption{Semantic evaluation on the Semeval dataset. (a) Quality evaluation of methods based on the word vector representation. (b) Close-up view.}
\label{fig:semantic-eval-Sem}
\end{figure}

Figures \ref{fig:eval-based-on-vectors-krpvn}, \ref{fig:eval-based-on-vectors-krpvn-zoom} and \ref{fig:eval-based-on-vectors-smvl}, \ref{fig:eval-based-on-vectors-smvl-zoom} show the results of the Krapivin and Semeval articles, respectively. In both datasets, the mean vector of RVA's keyphrases achieves higher cosine similarity (RVA) than the reference vector (Summary) and the mean vector of the low ranked candidate n-grams (Unselected). Furthermore, the four mean vectors of the other methods have lower cosine similarities in the majority of the articles than RVA, except for TfIdf-ft in the Semeval where the similarity values are at the same levels. Generally, the results based on the $F_1$-measure are consistent with the results of the semantic evaluation. The 50\% of the mean vectors that are included in the ``box'' part of the plot, i.e., from the first ($Q_1$) to the third ($Q_3$) quartile seem to interpret the methods' ranking according to the $F_1$-measure which goes almost hand in hand with the semantic evaluation. Even the slight superiority in the Semeval dataset of RVA over TfIdf-ft seems to be determined by this ``box'' part of the plot. Particularly, for RVA's box plot the $Q_1$ is equal to 0.9619 and the $Q_3$ is 0.9930, whereas for TfIdf the corresponding values are 0.9611 and 0.9924, respectively.

\subsection{Qualitative Results: RVA in Practice}
\label{quality_results}

In this section, we use RVA to extract the keyphrases of a publication based only on its title and abstract. This scientific article belongs to the Krapivin data collection. We quote its content below:
\\
\\
{\centering\fbox{\begin{minipage}{34em} \scriptsize{
Title: Clustering for Approximate Similarity Search in High-Dimensional Spaces.

Abstract: In this paper, we present a clustering and indexing paradigm (called Clindex) for high-dimensional search spaces. The scheme is designed for approximate similarity searches, where one would like to find many of the data points near a target point, but where one can tolerate missing a few near points. For such searches, our scheme can find near points with high recall in very few IOs and perform significantly better than other approaches. Our scheme is based on finding clusters and, then, building a simple but efficient index for them. We analyze the trade-offs involved in clustering and building such an index structure, and present extensive experimental results.
}
\end{minipage}}}
\\
\\
The corresponding set of the ``gold'' keyphrases are: 
\{\textit{clustering, approximate search, high-dimensional index, similarity search}\}. For evaluation purposes, we transform the set of ``gold'' keyphrases into the following one (stemmed keyphrases):

$\begin{aligned}
\{(cluster), (approxim, search), (highdimension, index) (similar, search)\}
\end{aligned}$

The RVA's result set is given in the first box below, followed by its stemmed version in the second box. The candidate keyphrases are presented by descending cosine similarity score. The words that are both in the golden set and in the set of our candidates are highlighted with bold typeface:\\

{\centering\fbox{\begin{minipage}{34em} { \{approximate similarity search, data points near, index structure, similarity search, approximate similarity, high recall, near points, points near, data points, finding clusters, search spaces, highdimensional search spaces, efficient index, target point, approximate similarity searches, near, indexing, search, highdimensional search, structure, present, data, clustering, recall\}
}
\end{minipage}}}

{\centering\fbox{\begin{minipage}{34em} { \{(\textbf{approxim}, \textbf{similar}, \textbf{search}), (data, point, near), (\textbf{index}, structur), (\textbf{similar}, \textbf{search}), (\textbf{approxim}, \textbf{similar}), (high, recal), (near, point), (point, near), (data, point), (find, \textbf{cluster}), (\textbf{search}, space), (\textbf{highdimension}, search, space), (effici, \textbf{index}), (target, point), (near), (\textbf{index}), (\textbf{search}), (\textbf{highdimension}, \textbf{search}), (structur), (present), (data), (\textbf{cluster}), (recal)\}
}
\end{minipage}}}
\\\\

The set of the returned keyphrases include all the words (unigrams) appearing in the ``gold'' keyphrases as well as additional keywords that have quite a strong role in the central meaning of the text. Furthermore, we notice that the RVA output is a set of keyphrases that are quite similar to each other, using a quite limited set of words: \{index, search, recal, target, point, similar, approxim, space, highdimension, structur, high, cluster, near, effici, data, find, present\}.

\begin{figure}[h]
\centering
\begin{subfigure}{.5\linewidth}
\centering
\includegraphics[width=1.0\linewidth]{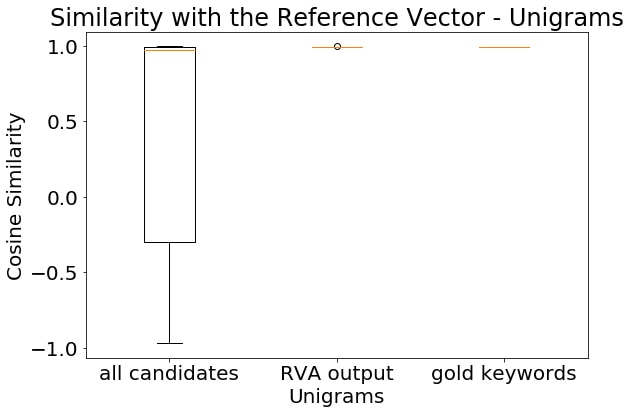}
\caption{}
\label{fig:uni}
\end{subfigure}%
\begin{subfigure}{.5\linewidth}
\centering
\includegraphics[width=1.0\linewidth]{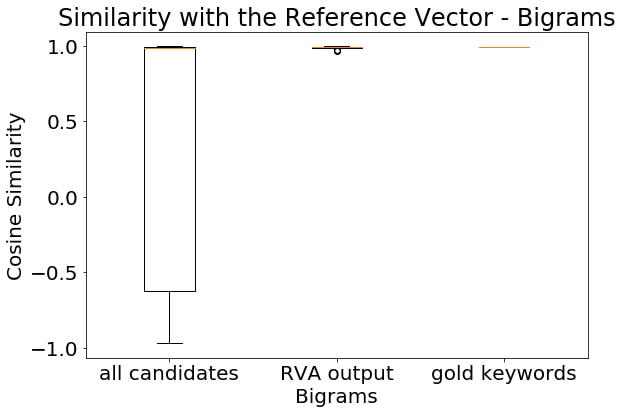}
\caption{}
\label{fig:bi}
\end{subfigure}\\[1ex]
\begin{subfigure}{.5\linewidth}
\centering
\includegraphics[width=1.0\linewidth]{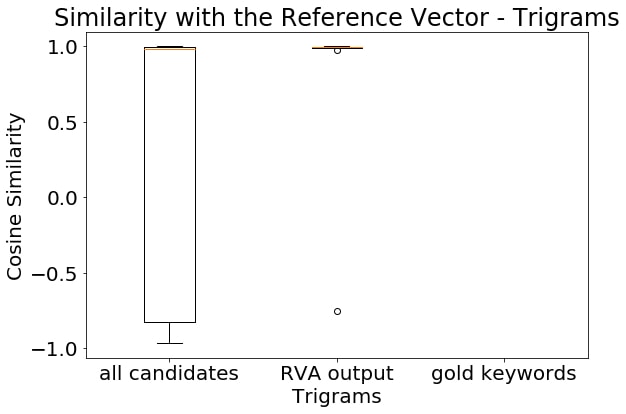}
\caption{}
\label{fig:tri}
\end{subfigure}%
\begin{subfigure}{.5\linewidth}
\centering
\includegraphics[width=1.0\linewidth]{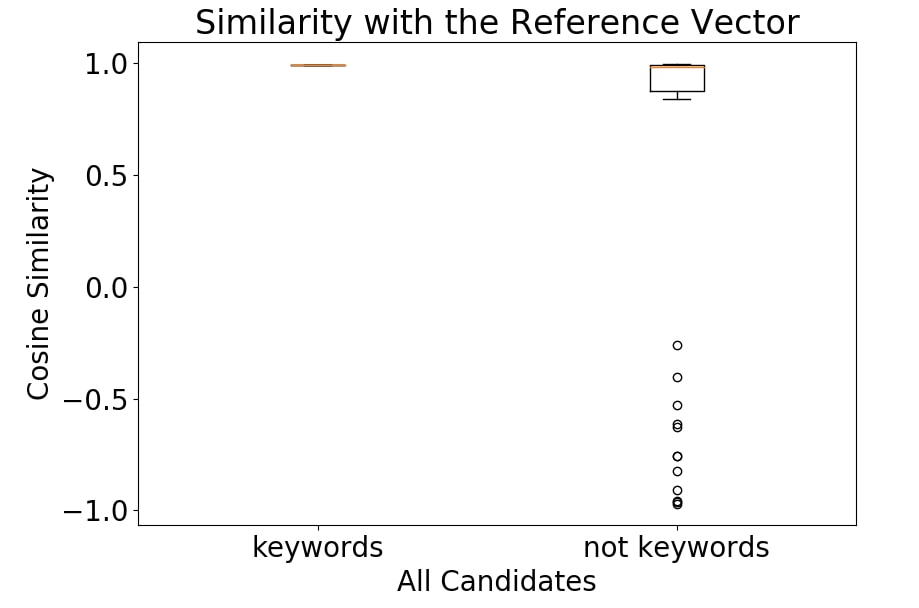}
\caption{}
\label{fig:example}
\end{subfigure}\\[1ex]
\caption{Boxplots of the extracted ngrams, (a) unigrams, (b) bigrams, (c) trigrams and their similarities to the reference vector. Figure (d) shows the similarity scores of all candidates as unigrams, grouped in ``Keywords'' and ``not keywords''}
\label{fig:ngrams}
\end{figure}

We also give two box plots in Fig. \ref{fig:example}, which summarize in a simple way that the ``gold'' keywords (bigrams and trigrams are also flattened as unigrams) that appear in the candidates accumulate at high cosine similarity values.
On the contrary, candidates that are not keywords cover a great range of cosine similarity values, as those words are not so close to the mean vector that is affected by words with a critical role (keywords). 


Moreover, in Fig. \ref{fig:ngrams}, we give the box plots of the extracted candidate n-grams, the RVA's output n-grams, and the ``gold'' keywords' n-grams with their similarities to the reference vector, in a more thorough view. We provide 3 separate figures (\ref{fig:uni}, \ref{fig:bi}, \ref{fig:tri}) to present the similarity of the unigrams, bigrams, and trigrams with the reference vector. We see that all types of n-grams have an expected behavior; RVA's n-grams are quite close to those of ``gold'' n-grams. In this way, we also confirm that the sum is a quite appropriate scoring function for bigrams and trigrams, as the corresponding similarities of the bigrams' and trigrams' mean vectors with the reference vectors are quite high, too.

For comparison reasons, we give the output produced by the baseline and the graph-based methods to provide a view of their keyphrases' quality.  In the previous section, we saw that it is better to use the articles’ summaries instead of the full-text for the keyphrase extraction task in order to avoid the noise of the full-text. For this reason, we present here, the best results of the graph-based algorithms (TopicRank-ab, SingleRank-ab, and WARank-ab) and the baseline (TfIdf-ab).

{\centering\fbox{\begin{minipage}{34em} { \textit{TopicRank-ab}:
\{(scheme), (\textbf{cluster}), (\textbf{approxim}, \textbf{similar}, \textbf{search}), (near, point), (\textbf{highdimension}, \textbf{search}, space), (clindex), (paradigm), (high, recal), (data, point), (effici, \textbf{index}), (simpl), (\textbf{index}, structur), (mani), (build), (target, point), (\textbf{highdimension}, space), (paper), (tradeoff), (better), (present, extens, experiment, result), (point)\}
}
\end{minipage}}}
\\\\
{\centering\fbox{\begin{minipage}{34em} { \textit{SingleRank-ab}:
\{(\textbf{highdimension}, \textbf{search}, space), (\textbf{approxim}, \textbf{similar}, \textbf{search}), (present, extens, experiment, result), (near, point), (\textbf{highdimension}, space), (\textbf{index}, structur), (data, point), (target, point), (effici, \textbf{index}), (point), (scheme), (high, recal), (\textbf{cluster}), (build), (tradeoff), (better), (mani), (paradigm), (clindex), (paper), (simpl)\}
}
\end{minipage}}}
\\\\
{\centering\fbox{\begin{minipage}{34em} { \textit{TfIdf-ab}:
\{(\textbf{approxim}, \textbf{similar}, \textbf{search}), (near, point), (\textbf{approxim}, \textbf{similar}), (\textbf{similar}, \textbf{search}), (\textbf{highdimension}), (clindex), (\textbf{index}, paradigm), (\textbf{highdimension}, \textbf{search}, space), (\textbf{highdimension}, \textbf{search}), (call, clindex), (toler, miss), (data, point, near), (present, extens, experiment), (\textbf{cluster}), (find, \textbf{cluster}), (extens, experiment, result), (high, recal), (tradeoff, involv), (target, point), (effici, \textbf{index}), (\textbf{highdimension}, space), (io), (present, extens), (point, near)\}
}
\end{minipage}}}
\\\\
{\centering\fbox{\begin{minipage}{34em} { \textit{WARank-ab}:
\{(\textbf{highdimension}, \textbf{search}, space), (\textbf{approxim}, \textbf{similar}, \textbf{search}), (such, \textbf{search}), (near, point), (target, point), (data, point), (point), (\textbf{highdimension}, space), (present, extens, experiment, result), (scheme), (few, io), (\textbf{index}, structur), (effici, \textbf{index}), (high, recal), (few), (build), (other, approach), (clindex), (t), (tradeoff), (better), (mani), (simpl), (\textbf{cluster})}\}
\end{minipage}}}
\\\\

The corresponding sets of unigrams that create the bigrams and the trigrams given above for each one method are presented below:

{\centering\fbox{\begin{minipage}{34em} { \textit{TopicRank-keywords}:
\textit{\{effici, point, approxim, high, cluster, paper, result, index, space, better, experiment, build, scheme, simpl, recal, highdimension, extens, data, present, clindex, search, target, structur, tradeoff, near, mani, paradigm, similar\}}
}
\end{minipage}}}
\\\\
{\centering\fbox{\begin{minipage}{34em} { \textit{SingleRank-keywords}:
\textit{\{effici, point, approxim, high, cluster, paper, result, index, space, better, experiment, build, scheme, simpl, recal, highdimension, extens, data, present, clindex, search, target, structur, tradeoff, near, mani, paradigm, similar\}}
}
\end{minipage}}}
\\\\
{\centering\fbox{\begin{minipage}{34em} { \textit{TfIdf-keywords}:
\textit{\{effici, point, approxim, high, cluster, result, io, miss, find, involv, index, space, experiment, call, recal, highdimension, extens, toler, data, present, clindex, search, target, tradeoff, near, paradigm, similar\}}
}
\end{minipage}}}
\\\\
{\centering\fbox{\begin{minipage}{34em} { \textit{WARank-keywords}:
\textit{\{near, point, data, target, approxim, similar, search, such, highdimension, space, present, extens, experiment, result, scheme, few, io, index, structur, effici, other, approach, clindex, build, high, recal, mani, better, simpl, cluster, t, tradeoff}\}}
\end{minipage}}}
\\\\

We see that the number of words involved in the formation of the keyphrases returned by TopicRank, SingleRank, WARank, and TfIdf algorithms is 28, 28, 32, 27, respectively, whereas for RVA the number of words is equal to 17, i.e., RVA's keyphrases revolve around a very specific and limited number of words instead of including additional redundant and irrelevant words like the other methods. 

\section{Conclusions and Future Work}
\label{contributions}


This work presented a new unsupervised keyphrase extraction method, whose main innovation is the use of {\em local} word embeddings, in particular GloVe vectors, to represent candidate keyphrases. Our empirical study offered evidence that such a representation can lead to better keyphrase extraction results, compared to using global representations, either pretrained on large corpora or focused on a given target corpus, as well as compared to popular state-of-the-art unsupervised keyphrase extraction approaches. 

We hope this work inspires other researchers to further investigate this novel local perspective of word embeddings. In particular we envisage implications of our work towards improved keyphrase extraction methods based on local word embeddings, towards applying local word embeddings to other information processing tasks, and towards developing novel methods for learning local word embeddings. 

In the near future we intend to build on top of this work, and develop graph-based unsupervised keyphrase extraction methods as well as supervised keyphrase extraction methods that rely on local GloVe vectors. We would also like to develop a solution that manages to extract keyphrases from the full-text of academic publications without being affected by the noise and redundancy that it contains.

\section*{Acknowledgements}

This work was partially funded by Atypon Systems Inc\footnote{\url{https://www.atypon.com/}}. 

\clearpage

\section*{References}

\bibliography{mybibfile}

\end{document}